\documentclass[lnicst]{svmultln}
\usepackage{url}
\usepackage{latexsym}
\usepackage{tipa}
\usepackage{graphicx}
\usepackage{makeidx}  
\usepackage{setspace}
\usepackage{enumitem}

\usepackage{float}
\floatstyle{ruled}
\newfloat{entry}{thp}{lop}
\floatname{entry}{Entry}

\usepackage[english]{babel}
\usepackage{ethiop}
\newcommand{\eth}{\selectlanguage{ethiop}}

\begin{document}
\selectlanguage{english}
\mainmatter              
\title{Minimal Dependency Translation:\\
a Framework for Computer-Assisted Translation\\
for Under-Resourced Languages}
\titlerunning{MDT}  
%
\author{Michael Gasser}
\authorrunning{Michael Gasser}   
%
\tocauthor{Michael Gasser}
\institute{Indiana University, Bloomington IN USA,\\
\email{gasser@indiana.edu},\\ WWW home page:
\texttt{http://homes.soic.indiana.edu/gasser/}
}

\maketitle              

\begin{abstract}
This paper introduces Minimal Dependency Translation (MDT), an ongoing project to develop a rule-based framework for the creation of rudimentary bilingual lexicon-grammars for machine translation and computer-assisted translation into and out of under-resourced languages as well as initial steps towards an implementation of MDT for English-to-Amharic translation. The basic units in MDT, called \textbf{groups}, are headed multi-item sequences. In addition to wordforms, groups may contain lexemes, syntactic-semantic categories, and grammatical features. Each group is associated with one or more translations, each of which is a group in a target language. During translation, constraint satisfaction is used to select a set of source-language groups for the input sentence and to sequence the words in the associated target-language groups.

\end{abstract}

\section{Introduction}
\label{intro}

For the majority of the world's languages we lack adequate resources to make use of the machine learning techniques that
have become the standard for modern computational linguistics.
For machine translation (MT) and computer-assisted translation (CAT),
the lack is even more serious because what is
required for machine learning is sentence-aligned translations,
which are even less common than monolingual corpora.
However, linguistic descriptions and sizable communities of native
speakers do exist for many under-resourced languages, including
Asian languages such as Telugu and Burmese, African languages such as
Amharic and Hausa, and indigenous American languages such as
Quechua and Guarani.
There is thus a need for frameworks that facilitate the rapid creation
of computational grammars and lexica by people and their automatic extension
through the limited corpora that are available.

The lack of computational linguistic resources for a language usually correlates with
a lack of written material in the language, an even more serious
disadvantage for the community of speakers.
The gap in available material is easily seen on Wikipedia, where the
Amharic edition currently has 13,767 articles and the Hausa edition
1,504 articles.
Compare these numbers with the editions for more privileged languages
with comparable numbers of speakers:
1,909,454 articles for Dutch and 1,237,519 for Polish.\cite{wikipedia}
One way to alleviate this gap between the more privileged and
less privileged languages is to accelerate the translation of documents
into the under-resourced languages.

For this reason, this project focuses on MT, and especially CAT, into languages
like Amharic.
The long-term goal is a system that allows users with little or no linguistic experience
to write bilingual lexicon-grammars
for low-resource languages that can also be updated on the basis of corpora,
when these are available, and that can be easily integrated into a CAT system, where
they are also updated on the basis of feedback from users.

This paper describes Minimal Dependency Translation (MDT), a lexical-grammatical framework for MT and CAT.
The core of MDT is a lexicon of phrasal units called groups.
A group's entry specifies translations to groups in one or more other languages.
Our focus to date has been on the language pairs Spanish-Guarani and
English-Amharic.
Examples and implementation details discussed in this paper are from
the English-Amharic system, called \textsc{Mit'mit'a}.

\section{Lexica and grammars}

\subsection{Phrasal lexica}
\label{subsect:phrase}

The idea of treating phrases rather than individual words as the basic units of a language
goes back at least to the proposal of a Phrasal Lexicon by Becker \cite{becker}.
In recent years, the idea has gained currency within the related frameworks of Construction Grammar \cite{steels}
and Frame Semantics \cite{fillmoreFS} as well as in phrase-based statistical machine translation (PBSMT).
Arguments in favor of phrasal units are often framed in terms of the ubiquity of idiomaticity, that is, departure
from strict compositionality.
Seen another way, phrasal units address the ubiquity of lexical ambiguity.
If a verb's interpretation depends on its object or subject, then it may make more sense to treat the combination
of the verb and particular objects or subjects as units in their own right.

Arguments based on idiomaticity and ambiguity are semantic, but they extend naturally to translation.
If the meaning of a source-language phrase fails to be the strict combination of the meanings
of the words in the phrase, then it is unlikely that the translation of the phrase will be the
combination of the translations of the source-language words.
Adding lexical context to an ambiguous word may permit an MT
system to select the appropriate translation.

\subsection{A simple phrasal lexicon}
\label{subsect:lexicon}

The basic lexical entries of MDT are multi-word units called \textbf{groups}.
Each group represents a catena \cite{osborneetal12}.
Catenae go beyond constituents, including all combinations of elements that are continuous
in the vertical dimension within a dependency tree.
For example, in the sentence \textit{I gave her a piece of my mind}, \{\textit{I,gave}\} and \{\textit{gave,her,piece}\}
are catenae but not constituents of the sentence.

A catena has a head, and each MDT group must also have a head, which indexes the group within the lexicon.
The other elements in the group are dependents of the head, but the
group has no further structure;
it is thus a {\em minimal} dependency structure.
A group's entry also specifies translations to groups in one or more other languages.
For each translation, the group's entry gives an \textbf{alignment}, representing inter-group correspondences between
elements, as in the phrase tables of PBSMT.
Entry~\ref{entry:way} shows a simple group entry of this sort.
The English group \texttt{$<$one way or the other$>$} with head \texttt{way}\footnote{In the figures, heads are enclosed in brackets.}
has as its Amharic translation the group
$<${\eth bazihm hona baziyA}$>$ (\textipa{\textit{bEzzih1m honE bEzziya}}),
which has its own entry in the Amharic lexicon.
In the alignment, three of the words in the English group are associated
with positions in the Amharic group; the others (indicated with ``0'') correspond to no word
in the Amharic group.

\begin{entry}
~\\
\small
\texttt{$<$one [way] or the other$>$}
\begin{itemize}[noitemsep]
   \item[] $\rightarrow$\texttt{amh} {\eth bazihm hona baziyA}
  \begin{itemize}[noitemsep]
  \item[] \texttt{align:[0,1,2,0,3]}
\end{itemize}
\end{itemize}
\normalsize
\caption{Group entry for \textit{one way or the other} and its Amharic translation}
\label{entry:way}
\end{entry}

\subsection{The lexicon-grammar tradeoff}
\label{subsect:lexgram}

A rudimentary lexicon with entries of this sort is simple in two senses: given an appropriate interface,
a user with no formal knowledge of linguistics can add entries in a
straightforward manner, and the resulting entries are easily understood.
Such a lexicon permits the translation of sentences that are combinations
of the wordforms in the group entries, as long as group order is preserved across
the languages and there are no constraints between groups that would affect the form
of the target-language words.
However, such a lexicon permits no
{\em generalization} to combinations of wordforms that are not explicit in the lexicon.
It would require a group entry for every reasonably possible combination of
wordforms.

At the other extreme from this simple lexicon is a full-blown grammar that is driven by
the traditional linguistic concern
with parsimony:
every possible generalization must be ``captured''.
Although such a grammar has the advantage of compactness and of reflecting general principles
of linguistic structure, it is difficult
to write, to debug, and to understand, requiring significant knowledge of linguistics.

In the MDT project, the goal is a range of possibilities along the continuum from
purely lexical (and phrasal) to syntactic/grammatical, with the emphasis on ease of entry
creation and interpretation.

\subsection{Lexemes}
\label{subsect:lexeme}

We can achieve significant generalization over simple groups consisting of wordforms by
permitting lexemes in groups.
As an example, consider the English group \texttt{$<$lose\_v hope$>$}, where \texttt{lose\_v} is
the verb lexeme \textit{lose}.
In order to make such a group usable, the system requires knowledge of verb morphology, either in the form
of a morphological analyzer or a list of wordforms associated with each lexeme in the lexicon.
For example, the system needs to be able to recognize that \textit{loses} is the third person singular present tense form of the lexeme \textit{lose\_v}.
MDT assumes such a resource for the source language and in addition a part-of-speech tagger to reduce the syntactic and morphological ambiguity that can result when words are analyzed in isolation.

Because a source-language lexeme will normally be translated as a lexeme rather than a wordform,
the system also requires knowledge of target-language morphology,
specifically either a morphological generator or a list of wordform associated with each
combination of lexeme and grammatical features.
For example, the system needs to know that for the Amharic verb {\eth .tafA} (\textipa{\textit{t'Effa}})\footnote{For simplification, Amharic verb lexemes are given in their usual citation form, the third person singular masculine perfect form. In fact they are represented internally in terms of their abstract roots, the sequence of consonants that characterize words in Semitic languages.}, one translation of \textit{lose}, the forms corresponding to \textit{loses} are {\eth \textsl{y.tafAl}} (\textipa{\textit{y1t'Efal}}) and {\eth \textsl{t.tafAla^c}} (\textipa{\textit{t1t'EfalE\v{c}\v{c}}}), for masculine and feminine respectively.
Entry~\ref{entry:lose1} shows the entry for the expression \textit{lose
hope} and its Amharic translation {\eth \textsl{tasfA qora.ta}} (\textipa{\textit{tEsfa k'orrEt'E}}),
literally `cut hope'.

\begin{entry}
~\\
\small
\texttt{$<$[lose\_v] hope$>$}
   \begin{itemize}[noitemsep]
   \item[]
   $\rightarrow$\texttt{amh} {\eth tasfA qora.ta}
   \begin{itemize}[noitemsep]
   \item[]
   \texttt{align:[2,1]} ; \texttt{agr:[([2,1],(tam:tam,sb:sb))]}
   \end{itemize}
   \end{itemize}
\normalsize
\caption{Group entry for \textit{lose hope} and its Amharic translation}
\label{entry:lose1}
\end{entry}

Because this entry accommodates multiple sequences of English wordforms,
we need to map these onto appropriate target-language sequences.
This is accomplished through pairs of agreement features
for the lexeme, constraining the corresponding target language form to agree with the source
form on those features.
In the example, the head \texttt{lose\_v} and its translation in the Amharic group agree on
the tense-aspect-modality (\texttt{tam}) and subject (\texttt{sb}) features.
For example, if this group is selected in the translation of the sentence \textit{John loses hope},
the head of the corresponding Amharic group will be constrained to be
third person singular present tense: {\eth \textsl{^gon tasfA yqor.tAl}} (\textipa{\textit{\v{\j}on tEsfa y1k'ort'al}}).\footnote{In fact the system would also generate the (incorrect) feminine of the verb in this case since the group does not include the subject itself: {\eth \textsl{tqor.talA^c}}.}
For more about where the features of source-language words come from, see the section on morphosyntactic transformations below.

\subsection{Lexical/grammatical categories}
\label{subsect:cats}

Another straightforward way to generalize across groups is to introduce syntactic or semantic categories.
Consider the English expression \textit{make fun of somebody}.
We can generalize across specific word sequences such as \textit{made fun of her}
and \textit{made fun of the mayor} by replacing the specific wordforms in position
4 in the group with a category that includes the wordforms that can fill that position.
This requires a dictionary of category labels for wordforms.
Entry~\ref{entry:fun} shows how this appears in the lexicon.
Category names are preceded by \$.

\begin{entry}
~\\
\small
\texttt{$<$[make\_v] fun of \$sbd$>$}
   \begin{itemize}[noitemsep]
   \item[]
   \texttt{$\rightarrow$ amh \$sbd[+acc]} {\eth 'a^sofa}
   \begin{itemize}[noitemsep]
   \item[]
   \texttt{align:[2,0,0,1]} ; \texttt{agr:[([2,1],(tns:tns,sb:sb)),([4,1],(num:num))]}
   \end{itemize}
   \end{itemize}
\normalsize
\caption{Group entry for \textit{make fun of somebody} and its Amharic translation}
\label{entry:fun}
\end{entry}

\begin{figure}[ht]
\begin{center}
  \includegraphics{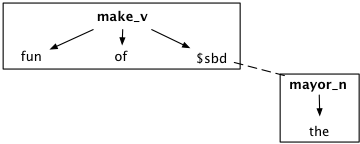}
\caption{Merging of two groups in \textit{make fun of the mayor}}
\label{fig:fun}
\end{center}
\end{figure}

Because group positions that are filled by categories do not specify a surface form,
during translation they must be merged with other groups that match
the category and do specify a form.
For example, in the translation of the sequence \textit{made fun of the mayor},
position 4 in the group \texttt{$<$make\_v fun of \$sbd$>$} may be
filled by the head of the group \texttt{$<$the mayor$>$}.
This \textbf{node merging} process is illustrated in Figure~\ref{fig:fun}.

Finally, the Amharic group that is the translation of \textit{make fun of \$sbd} includes the
constraint that whatever word fills the role of \textit{\$sbd} in the Amharic sentence must be accusative
since it is the direct object of the verb {\eth 'a^sofa}.


\subsection{Morphosyntactic transformations}
\label{subsect:morphosyn}

For languages pairs, such as English and Amharic, that differ greatly in their syntax and morphology,
a further elaboration of the framework permits many generalizations that save on the number of
groups required.
Amharic verb morphology is extremely complex, including tense-aspect-modality, subject and sometimes object agreement, as well as morphemes indicating whether the verb is negative and the main verb of a relative clause.
In English almost all of these features are indicated syntactically rather than morphologically.
Consider the sentence \textit{they do not know her} and its Amharic translation {\eth 'ayAwquAtm} (\textipa{\textit{ayawk\super w'at1m}}).
With a different subject, different object, or an affirmative rather than a negative verb, the single Amharic verb translating this sentence would differ.
Clearly without including the subject, the object, or the word \textit{not} in groups, there is no way the Amharic morphology can end up correct.
However, doing this entails a significant combinatorial explosion: for each English verb there would need to be hundreds of groups to cover
all the different combinations of subject, object, polarity, and tense-aspect-modality.

We can deal with some aspects of this problem by incorporating a pre-processing phase that modifies the English input to make it more like Amharic.
Following morphological analysis of the input sentence, it is matched against a set of morphosyntactic transformation rules,
each of which may delete words and/or modify the features of words.
One of these is shown in Entry~\ref{entry:theyneg}.
This rule matches sequences in input sentences consisting of the words \textit{they do not} followed by a verb.
It modifies the features of the verb to make its subject third person plural, its aspect imperfective (corresponding roughly to English present and future), and its
polarity negative, and it deletes the first three words of the sequence.

\begin{entry}
~\\
\small
\texttt{they do not \$v[sb$=$3p,tam$=$impf,$+$neg]} ; \texttt{del 1, 2, 3}
\normalsize
\caption{Morphosyntactic transformation rule for \textit{they do not \$v}}
\label{entry:theyneg}
\end{entry}

Getting all of this morphology right is a challenge, and the MDT approach only handles some of the cases.
Since the morphosyntactic transformation phase does not actually involve a syntactic parse of the input sentence, gaps present a problem.
For example, adverbs can intervene between \textit{not} and the main verb in English, in which case a rule such as that in Entry~\ref{entry:theyneg}
would not apply.

\section{Constraint satisfaction and translation}
\label{sect:cs}

The steps in MDT translation are illustrated in Figure~\ref{fig:fun2} for the input sentence \textit{she made fun of the mayor}.

\begin{figure}[ht]
\small
\texttt{(1) She made fun of the mayor.}\\
\texttt{(2) she make\_v[tns=pst] fun\_n of the mayor\_n}\\
\texttt{(3) make\_v[tns=pst,tam=prf,sb=3psf] fun\_n of mayor\_n[+def]}\\
\texttt{(4) $<$make\_v fun of \$sbd$> <$mayor\_n$>$}\\
\texttt{(5) $<$make\_v[tns=pst,tam=prf,sb=3psf] fun of $<$mayor\_n[+def]$>>$}\\
\texttt{(6)} $<<${\eth kantibA}\texttt{\_n[+def,+acc]$>$} {\eth 'a^sofa}\texttt{\_v[tam=prf,sb=3psf]$>$}\\
\texttt{(7)} {\eth kantibAwn 'a^sofa^c}
\normalsize
\caption{Steps in the translation of \textit{she made fun of the mayor}}
\label{fig:fun2}
\end{figure}

Following tokenization of the input sentence, the wordforms in the sentence are tagged for part-of-speech and analyzed morphologically (2).
Next, the sequence of analyzed words is matched against the morphosyntactic transformation rules (3).
In the example sentence two rules match, one for \textit{she} followed by a past tense verb, one for \textit{the} followed by a noun.
The first rule assigns perfective aspect to the verb and deletes \textit{she}.
The second assigns definiteness to the noun and deletes \textit{the} (definiteness is expressed by a suffix in Amharic).
Next the words or lexemes resulting from this first pass are used to look up candidate groups in the
groups dictionary (4).

To complete sentence analysis, the system assigns a set of groups to the input sentence.
A successful group assignment associates as many words in the sentence as possible with a group,
and no word to more than one group, unless that word represents a node merging (see below).
Longer groups over sequences of shorter groups.
Node merging takes place during this phase; in the example, the \$sbd node in the instantiation of the group
\texttt{$<$make\_v fun of \$sbd$>$} is merged with an instantiation of the group \texttt{$<$mayor\_n$>$} (5).

Group selection is implemented in the form of constraint satisfaction, making use of insights
from the Extensive Dependency Grammar framework (XDG) \cite{debusmann}.
Although considerable source-sentence ambiguity is eliminated because groups
incorporate context, ambiguity is still possible, particularly for figurative expressions
that also have a literal interpretation.
In this case, the constraint satisfaction process undertakes a search through the space of possible group
assignments, creating an analysis for each successful assignment.

During the transfer phase, a source-language group assignment is converted to an assignment of the crorresponding target-language groups (6).
In this process some target-language items are assigned grammatical features on the basis of
cross-language agreement constraints from the source group's entry.
In the example sentence, the Amharic verb gets its \texttt{tam} and \texttt{sb} feature values
from the English verb, and the noun gets its \texttt{def} feature value from the English noun.
A source-language group may have more than one translation; unless specified otherwise, the
transfer phase returns all of these.

During the realization phase,
surface forms are generated for each target-language group assignment, based on the lexemes and grammatical features that resulted from
the transfer phase (7).

\section{Related work}
\label{sect:related}

Our goals are similar to those of the Apertium \cite{apertium} project.
As with Apertium, we are developing open-source, rule-based systems for MT, and
we work within the framework of relatively shallow, chunking grammars.
We differ mainly in our willingness to sacrifice linguistic coverage
to achieve our goals of flexibility, robustness, and transparency.
We accommodate a range of lexical-grammatical possibilities, from the completely
lexical on the one extreme to phrasal units consisting of a single lexeme and one or more syntactic/semantic
categories on the other, and we are not so concerned that MDT
grammars will accept many ungrammatical source-language
sentences or even output ungrammatical (along with grammatical)
translations.
Because MDT focuses on the translation of phrases and outputs
usually outputs multiple translations rather than complete
sentences, it is more appropriate for CAT than for full-scale MT.

In terms of long-term goals, MDT also resembles the Expedition
project \cite{mcshane+nirenburg}, which makes use of knowledge acquisition
techniques and naive monolingual informants to develop
MT systems that translate low-resource languages into English.
Our project differs first, in assuming bilingual informants and second, in aiming to
develop systems that are unrestricted with respect to target language.
In fact we are more interested in MT systems with low-resource languages as target languages
because of the lack of documents in such languages.

Although MDT is not intended as a linguistic theory, it is worth
mentioning which theories it has the most in common with.
Like Construction Grammar \cite{steels} and Frame Semantics \cite{fillmoreFS},
it treats linguistic knowledge as essentially phrasal.
Like synchronous context-free grammar (SCFG) \cite{chiang}, it associates multi-word units in
two languages, aligning the elements of the units and representing word order within each.
MDT differs from SCFG in having nothing like rewrite rules or non-terminals.
MDT belongs to the family of dependency grammar (DG) theories because the heads of its
phrasal units are words or lexemes rather than non-terminals.
However, it remains an extremely primitive form of DG,
permitting only flat structures with unlabeled arcs and no relations between groups
other than through the merge operation described in Section~\ref{subsect:cats}.
This means that complex grammatical phenomena such as long-distance dependencies and
word-order variability can only be captured through specific groups.

\section{English-Amharic implementation}
\label{sect:implementation}

We are in the process of creating an English--Amharic implementation of
MDT, called \textsc{Mit'mit'a} ({\eth mi.tmi.tA}).
In doing so, we have relied on the tokenizer and POS tagger from
spaCy\cite{spacy}, the Amharic-English dictionary
of Amsalu Aklilu\cite{amsalu}, the Amharic morphological generator within the HornMorpho system
for morphological processing\cite{gasser}, the extensive grammatical
descriptions of Amharic, and the author's own knowledge of the grammars of English and Amharic.
Although far from finalized, the implementation already contains
approximately 7000 groups and 500 morphosyntactic transformation
rules.
The MDT framework offers a range of possibilities with respect to how
many grammatical generalizations are captured through the use of
morphosyntactic transformations and category nodes in groups, and
\textsc{Mit'mit'a} falls on the heavily grammatical end of the spectrum.
Thus there are transformation rules accommodating combinations of
pronoun subjects with all English tenses and modal verbs in both
affirmative and negative forms.
The result is that the system often, though by no means always, gets
Amharic morphology right.

Evaluation of an MDT implementation should be of two types: for accuracy of the translations
and for usability of the system with CAT.
Since \textsc{Mit'mit'a} is still under development, we have not undertaken a systematic
evaluation of either of these sorts.
However, we have begun to informally compare the system's accuracy with
that of existing statistical MT systems.


There are several commercial English--Amharic machine translation
systems, including Google Translate \cite{google} and Abyssinica Translator \cite{ethiocloud}, developed
by EthioCloud.
We can highlight the strengths of \textsc{Mit'mit'a} by examining the
grammaticality of the output of Google Translate fpr the grammatical
patterns that \textsc{Mit'mit'a} is designed to capture.
\textsc{Mit'mit'a} knows how to translate all patterns consisting of a pronoun
subject followed by a negative or affirmative verb in any of eight
possible English tenses/aspects and combinations of modal and main
verbs, as well as combinations of transitive verbs with personal pronoun objects.
Given the roughly 2,700 English verbs that the sytem has Amharic
translations for, the result is hundreds of thousands of translatable
patterns such as \textit{she is about to break it}.
Randomly selecting from the possible verb patterns, pronouns, and six
common verbs, we get an idea of how well Google Translate performs
on such combinations.
Of the 54 resulting sentences, Google Translate outputs only one grammatically correct verb.
\textsc{Mit'mit'a}, on the other hand, makes only one minor mistake on three sentences, treating \textit{him} in \textit{write him} as a direct rather than an indirect object.
\textsc{Mit'mit'a} also has the advantage of returning multiple translations when there is ambiguity, for example, translating English \textit{you} in three ways (feminine singular, masculine singular, plural).
Needless to say, this is not really a fair comparison since it is based on examples that \textsc{Mit'mit'a} is designed to handle, but it does give an idea of what sorts of advantages this rule-based system can have over a statistical system for a morphologically complex language in the context of limited training data.

\section{Status of project, ongoing and future work}
\label{sect:status}

MDT code, including implementations for Spanish--Guarani and English--Amharic,
is available at \url{https://github.com/hltdi/mainumby} and \url{https://github.com/hltdi/mitmita}
under the GPL license.

In order to develop more complete lexicon-grammars for English--Amharic and
Spanish--Guarani,
we are working on methods for automatically extracting groups from the
limited bilingual corpora that are available.
We are also implementing an interface for the use of MDT implementations for CAT;
working versions for Spanish-Guarani and English-Amharic can be found at \url{https://plogs.soic.indiana.edu/mainumby/} and \url{https://plogs.soic.indiana.edu/mitmita/}.
Here it will be important to evaluate to what extent translators find their task simplified through the use of the system.
Finally, since the interface records the user's translations whether or not they make use
of the suggestions from MDT, there is the opportunity to update the system's lexicon-grammar
on the basis of those translations.



\section{Conclusions}
\label{sect:conclusions}

Relatively sophisticated computational grammars, parsers, and/or generators
exist for perhaps a dozen languages, and usable MT systems exist for
at most dozens of pairs of languages.
This leaves the great majority of languages and the communities who speak them
even more disadvantaged than they were before the digital revolution.
What is called for are methods that can be quickly and easily deployed to
begin to record the grammars and lexica of these languages and to use these
tools for the benefit of the linguistic communities.
The MDT project is designed with these needs in mind.
Though far from achieving our ultimate goals, we have developed a simple, flexible, and robust
framework for bilingual lexicon-grammars and MT/CAT that we hope will be a starting
point for a large number of under-resourced languages.


\begin{thebibliography}{14}

\bibitem{ethiocloud}
Abyssinica Translator:
\url{http://translator.abyssinica.com/}

\bibitem{amsalu}
Amsalu Aklilu: Amharic-English Dictionary.
Addis Ababa. Kuraz (1979)

\bibitem{becker}
Becker, J.: The Phrasal Lexicon.
In: Schank, R., Nash-Webber, B. (eds.) Theoretical Issues in Natural Language Processing, pp. 38--41.
Association for  Computational Linguistics (1975)


\bibitem{chiang}
Chiang, D.: Hierarchical Phrase-Based Translation.
Computational Linguistics 33, 201--228 (2007)

\bibitem{debusmann}
Debusmann, R.: Extensible Dependency Grammar: A Modular Grammar Formalism Based
On Multigraph Description.
Ph.D. thesis, Universit\"{a}t des Saarlandes (2007)

\bibitem{fillmoreFS}
Fillmore, C.J., Baker, C.F.: Frame Semantics for Text Understanding.
In: Proceedings of WordNet and Other Lexical Resources Workshop, NAACL (2001)

\bibitem{apertium}
Forcada, M.L,  Ginest\'{i}-Rosell, M., Nordfalk, J., O'Regan, J.,
Ortiz-Rojas, S., P\'{e}rez-Ortiz, ,J.A., S\'{a}nchez-Mart\'{i}nez, F.,
Ram\'{i}rez-S\'{a}nchez, G., Tyers, F.M.:
Apertium: a Free/Open-Source Platform for Rule-Based Machine Translation.
Machine Translation 25, 127--144 (2011)


\bibitem{gasser}
Gasser, M.: HornMorpho: a System for Morphological Processing of Amharic, Oromo, and Tigrinya.
In: Proceedings of Conference on Human Language Technology for Development, Alexandria, Egypt
(2011)

\bibitem{google}
Google Translate: English-Amharic, \url{https://translate.google.com/\#en/am/}

\bibitem{mcshane+nirenburg}
McShane, M., Nirenburg, S., Cowie, J., Zacharski, R.:
Embedding Knowledge Elicition and MT Systems Within a Single Architecture.
Machine Translation 17, 271--305 (2002)

\bibitem{osborneetal12}
Osborne, T., Putnam, M., Gross, T.:
Catenae: Introducing a Novel Unit of Syntactic Analysis.
Syntax 15, 354--396 (2012)

\bibitem{spacy}
Honnibal, M.:
spaCy: Industrial-Strength Natural Language Processing in Python.
\url{https://spacy.io/}
(2016)

\bibitem{steels}
Steels, L. (ed.):
Design Patterns in Fluid Construction Grammar.
John Benjamins, Amsterdam (2011)


\bibitem{wikipedia}
Wikipedia (English): List of Wikipedias (date retrieved, 29 August
2017).
\url{https://en.wikipedia.org/w/index.php?title=List_of_Wikipedias&oldid=796394102} (2017)

\end{thebibliography}

\end{document}